\documentclass{article}

\usepackage[preprint, nonatbib]{neurips_2024}

\usepackage[utf8]{inputenc} 
\usepackage[T1]{fontenc}    
\usepackage{hyperref}       
\usepackage{url}            
\usepackage{booktabs}       
\usepackage{amsfonts}       
\usepackage{nicefrac}       
\usepackage{microtype}      
\usepackage{xcolor}         
\usepackage{graphicx}  
\usepackage{amsmath}   
\usepackage{multirow}  

\title{Dynamic Context Adaptation and Information Flow Control in Transformers: Introducing the Evaluator Adjuster Unit and Gated Residual Connections}

\author{%
  Sahil Rajesh Dhayalkar
  \\
  Brain Corporation \\
  San Diego, CA\\
  \texttt{sahil.dhayalkar@braincorp.com} \\
}

\begin{document}

\maketitle

\begin{abstract}
  Transformers have revolutionized various domains of artificial intelligence due to their unique ability to model long-range dependencies in data. However, they lack in nuanced, context-dependent modulation of features and information flow. This paper introduces two significant enhancements to the transformer architecture—the Evaluator Adjuster Unit (EAU) and Gated Residual Connections (GRC)—designed to address these limitations. The EAU dynamically modulates attention outputs based on the relevance of the input context, allowing for more adaptive response patterns. Concurrently, the GRC modifies the transformer's residual connections through a gating mechanism that selectively controls the information flow, thereby enhancing the network's ability to focus on contextually important features. We evaluate the performance of these enhancements across several benchmarks in natural language processing. Our results demonstrate improved adaptability and efficiency, suggesting that these modifications could set new standards for designing flexible and context-aware transformer models.
\end{abstract}

\section{Introduction}
\label{Introduction}
The Transformer model, introduced in \cite{NIPS2017_3f5ee243}, has become the cornerstone of modern natural language processing (NLP) and is increasingly permeating other domains such as computer vision and audio processing. Its core mechanism, self-attention, allows it to capture long-range dependencies and handle sequences with remarkable effectiveness. However, as the adoption and adaptation of transformer architectures have grown, so too have the challenges associated with their computational efficiency, scalability, and ability to dynamically adapt to the nuanced requirements of varied tasks.

Recent advancements in machine learning have increasingly focused on enhancing the adaptability and efficiency of transformer architectures. Modifications to the foundational components of transformers, such as attention mechanisms and residual connections, have shown promising results in addressing these challenges. Despite these efforts, the quest for more adaptable and efficient models remains at the forefront of research, particularly in scenarios demanding dynamic context-aware processing.

In response to these challenges, this work introduces two novel enhancements to the transformer architecture: the Evaluator Adjuster Unit (EAU) and Gated Residual Connections (GRC). These modules are designed to improve the model's performance by enabling more dynamic and context-sensitive adjustments within the network. The EAU dynamically modulates attention outputs by assessing and adjusting the relevance of attention scores, thereby tailoring the network's responses based on the input context. Concurrently, the GRC enhances the transformer's residual connections by integrating a gating mechanism that controls the flow of information, allowing the model to selectively emphasize or suppress features based on their contextual importance.

This paper is structured as follows: following this introduction, we present a comprehensive background and literature survey that outlines previous efforts in Section \ref{Background} and sets the stage for our contributions. We then detail our proposed approach in Section \ref{Proposed_Approach}, including the theoretical foundation and implementation specifics of the EAU and GRC, and evaluate these enhancements across several benchmarks in Section \ref{Experiments_and_Evaluation}. Finally, we discuss the limitations of our findings in Section \ref{Limitations} before concluding in Section \ref{Conclusion}.

Our contributions are twofold: First, we propose the Evaluator Adjuster Unit, which introduces a novel method for context-dependent modulation of attention, enhancing the transformer's adaptability and responsiveness. Second, we develop Gated Residual Connections, which extend the transformer's capability to manage information flow through adaptive gating, potentially leading to more nuanced and effective processing. Together, these enhancements aim to set a new standard for the design of flexible and efficient transformer models.

\section{Background}
\label{Background}

The advent of transformer architectures has revolutionized the field of natural language processing (NLP) and beyond, primarily due to their ability to capture long-range dependencies and their scalability in handling large datasets. \cite{NIPS2017_3f5ee243} introduced the Transformer model, which eschews recurrent layers in favor of self-attention mechanisms, providing a new paradigm for sequence learning tasks.

However, despite their success, transformers are not without limitations. For instance, they can be computationally expensive and may struggle with context-dependent adjustments of features based on their relevance. This has led to significant research aimed at improving their efficiency and effectiveness. In particular, modifications to attention mechanisms and information flow within transformers have been a focal point.

\subsection{Adaptations in attention mechanisms}
Attention mechanisms, the core of transformer architectures, have seen various adaptations to enhance model performance and interpretability. \cite{kitaev2020reformer} proposed the Reformer, which reduces memory consumption by limiting the self-attention computation to a subset of key elements. A significant advancement, known as the Sparse Transformer [5], employs sparse factorizations of the attention matrix, enabling the model to handle longer sequences efficiently without a corresponding rise in computational demands. \cite{wang2020linformer} introduced Linformer, which projects the attention matrix into a lower-dimensional space, significantly reducing the computational complexity from quadratic to linear with respect to sequence length. This adaptation maintains performance while enhancing efficiency, making it suitable for longer sequences. \cite{choromanski2020rethinking} developed the Performer, which utilizes random feature maps through the Fast Attention Via positive Orthogonal Random features approach (FAVOR+) to approximate the softmax function in attention. This method allows the Performer to scale linearly in terms of memory and compute, irrespective of sequence length.

\subsection{Context-dependent modulation}
Efforts to allow transformers to adapt their behavior dynamically based on context have also emerged. The introduction of conditional computation within transformers, as explored in \cite{bengio2013estimating}, suggests mechanisms where parts of the network are activated conditionally based on the input, potentially increasing model efficiency and capacity for handling complex dependencies. \cite{shen2018bidirectional} explored an architecture where the scope and focus of the attention mechanism are modulated by additional contextual information from the rest of the network, thereby enhancing the relevance of attended features and improving performance on tasks requiring nuanced understanding. \cite{sukhbaatar-etal-2019-adaptive} proposed dynamically adjustable attention spans, where the extent of attention can be modified based on the task at hand, allowing models to either focus narrowly on important aspects or broadly to integrate wider contextual information.

\subsection{Residual connections}
The standard residual connections in Transformers \cite{NIPS2017_3f5ee243} facilitate training deep architectures by allowing gradients to flow through the networks more effectively. However, these connections are typically static and do not adapt to the context of the input data. Researchers have begun exploring adaptive or conditional residuals as a means to improve the representational power of models. For instance, \cite{Sabour10.5555/3294996.3295142} introduced capsules in neural networks that use dynamic routing between layers as a form of adaptive residuals, which could be seen as an inspiration for contextually gated connections.

\subsection{Gated mechanisms}
Gated mechanisms have been widely used in various architectures, like GRUs \cite{cho-etal-2014-learning} and LSTMs \cite{Hochreiter10.1162/neco.1997.9.8.1735}, to control the flow of information. They are particularly effective in recurrent setups but less explored in the context of transformers. \cite{Dauphin10.5555/3305381.3305478} utilized gating mechanisms within CNNs to control information flow, demonstrating their effectiveness in non-recurrent architectures as well.

\subsection{Proposed contributions}
This paper introduces two novel modules: the Evaluator Adjuster Unit (EAU) and Gated Residual Connections (GRC), designed to address these issues. The EAU dynamically modulates attention outputs, enhancing the transformer's adaptability and response to the input context, echoing the conditional computation paradigms suggested in \cite{bengio2013estimating} but applied directly within the transformer framework. Meanwhile, the GRC enhances the transformer’s residual connections by incorporating a gating mechanism that selectively emphasizes or suppresses information flow, thereby increasing the model’s capacity to manage information relevance effectively.

Both proposed enhancements aim to refine the transformer architecture’s capability to process and represent complex dependencies more efficiently. These contributions are poised to set a precedent for further explorations into making transformer models not only more computationally efficient but also contextually aware and adaptive.


\section{Proposed approach}
\label{Proposed_Approach}

In this work, we introduce two innovative neural network modules: the Evaluator Adjuster Unit and Gated Residual Connections. These modules are designed to enhance the adaptability and effectiveness of transformer-based architectures. The modules are straightforward and can be easily integrated into any transformer-based architectures.

\subsection{Evaluator Adjuster Unit}
The Evaluator Adjuster Unit (EAU) is a dual-component module designed to dynamically modulate attention outputs by first assessing the incoming attention scores and subsequently tailoring adjustments based on this assessment. It consists of an Evaluation network, which produces context-dependent scoring vectors, and an Adjustment network, which computes adaptive modifications.

\subsubsection{Evaluation network}
The Evaluation network generates a scoring vector that gauges the relevance of various components of the attention scores through the following transformations:
\begin{itemize}
    \item \textbf{Linear transformation and non-linearity:}
    Let $ \mathbf{x} \in \mathbb{R}^k $ be the input attention scores. $k$ is the dimension of key, query and value of the transformer. The input attention scores $\mathbf{x}$ undergoes a linear transformation, followed by a ReLU activation to introduce non-linearity:
    \begin{equation}
    \mathbf{h} = \text{ReLU}(\mathbf{W}_1 \mathbf{x} + \mathbf{b}_1) \\
    \end{equation}
    where $ \mathbf{W}_1 \in \mathbb{R}^{\frac{k}{2} \times k} $ and $ \mathbf{b}_1 \in \mathbb{R}^{\frac{k}{2}}$ are the trainable weights and biases of the first layer respectively. In our implementation, we reduce the dimensionality of the input attention by half (via $\mathbf{W}_1$ and $\mathbf{b}_1$) to allow for the Evaluation network to be light weight, while also potentially allowing to focus on capturing the most salient features.
    \item \textbf{Scoring vector via sigmoid activation:}
    The hidden representation $\mathbf{h}$ is further transformed to produce a scoring vector $\mathbf{e}$, constrained between 0 and 1 using a sigmoid $\sigma$ function:
    \begin{equation}
    \mathbf{e} = \sigma(\mathbf{W}_2 \mathbf{h} + \mathbf{b}_2)
    \end{equation}
    The scoring vector $\mathbf{e}$, constrained by $ \mathbf{W}_2 \in \mathbb{R}^{k \times \frac{k}{2}} $ and $ \mathbf{b}_2 \in \mathbb{R}^k$, provides interpretable importance scores. We upsample the output back to match the dimensions of input $\mathbf{x}$.
\end{itemize}

The Evaluation network outputs a vector of evaluation scores with the same size as $\mathbf{x}$. Each element of the evaluation scores indicates the relative importance or the quality of the corresponding element in $\mathbf{x}$.

\subsubsection{Adjustment network}
Simultaneously, the Adjustment network computes a vector of modifications $a$, which adjust the input based on the evaluations:
\begin{equation}
\mathbf{a} = \tanh(\mathbf{W}_3 \mathbf{x} + \mathbf{b}_3)
\end{equation}
where $ \mathbf{W}_3 \in \mathbb{R}^{k \times k} $ and $ \mathbf{b}_3 \in \mathbb{R}^k$ denote the weights and biases of the adjustment layer respectively. The $\tanh$ ensures that the adjustment factors are bounded between [-1, 1], which helps in keeping the adjusted values within a reasonable range, preventing drastic changes which could destabilize the learning process. 

\subsubsection{Output computation and integration}
\begin{figure}
  \centering
  \includegraphics[width=0.99\textwidth]{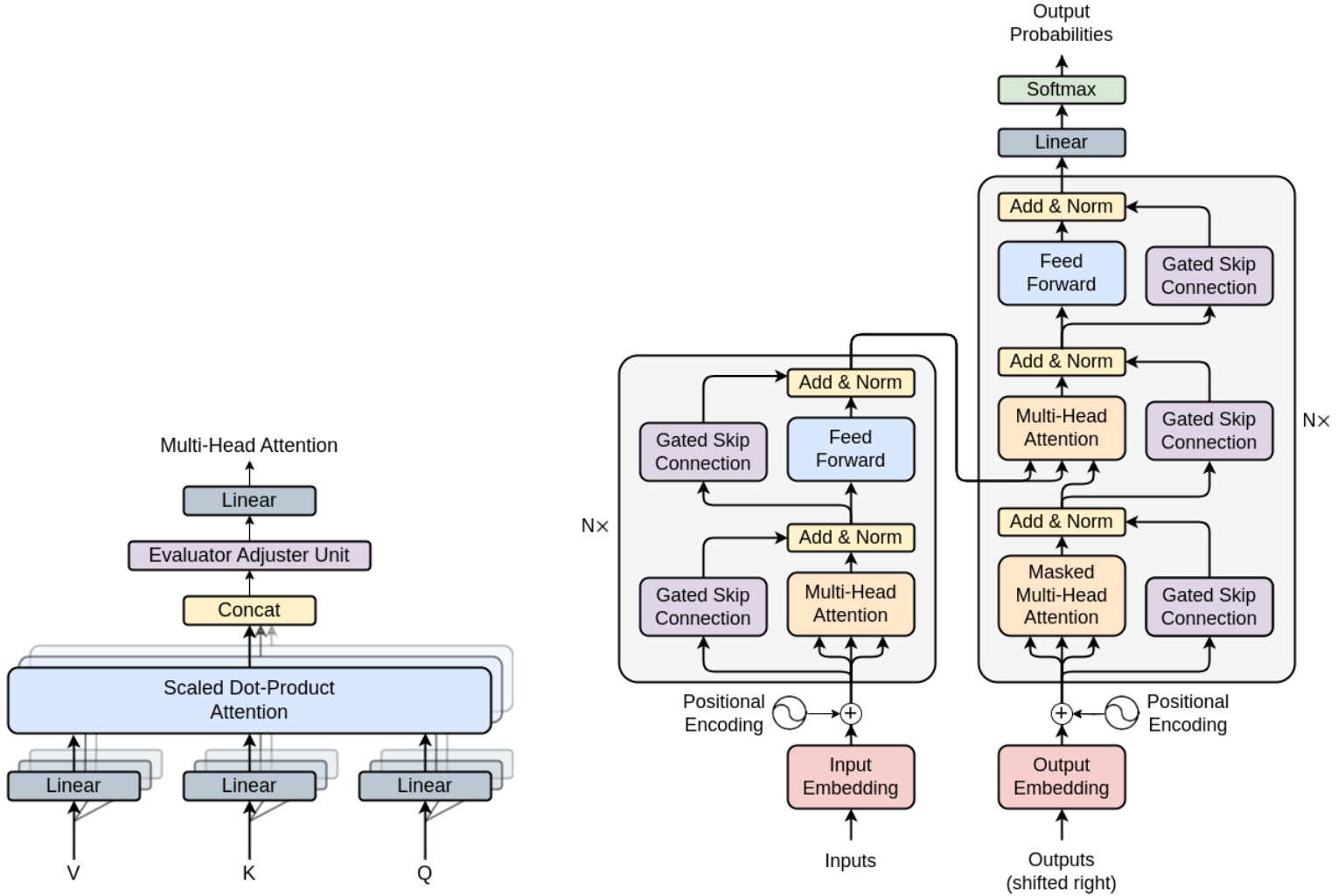}
  \caption{Integration of Evaluator Adjuster Unit (left) and Gated Residual Connections (right) in the Transformer model introduced in \cite{NIPS2017_3f5ee243}.}
  \label{integration_img}
\end{figure}

The outputs of both networks are integrated to dynamically adjust the original input:
\begin{equation}
\mathbf{y} = \mathbf{x} + (\mathbf{a} \odot \mathbf{e})
\end{equation}
The adjustment factors provided by the Adjustment network which are then element-wise multiplied by the evaluation scores provided by the Evaluation network, thus integrating the importance scores into the adjustment factors and modulating how much each element of the input should be adjusted. This operation allows for precise, context-aware adjustments, enhancing the model's ability to handle complex dependencies. After multi-head attention and before each feed-forward network in the encoder and decoder layers of the transformer architecture proposed in \cite{NIPS2017_3f5ee243}, the outputs are processed through an EAU, allowing dynamic adjustments based on the context provided by the attention mechanisms. Refer Figure \ref{integration_img} to visually see the integration of Evaluator Adjuster Units in the Transformer model introduced in \cite{NIPS2017_3f5ee243}.

\subsubsection{Intuition behind the Evaluator Adjustor Unit}
\begin{itemize}
    \item Dynamic feature modulation: By combining evaluation with adjustment, this unit dynamically modulates the features based on their evaluated importance. This could be particularly useful in scenarios where certain features need to be emphasized or suppressed based on the context provided by other parts of the model or input data.
    \item Self-adaptation: It allows the model to adapt its own outputs during training, potentially leading to more robust learning as the model can learn to focus more on important features and less on noise or irrelevant details.
    \item Enhanced representation: This mechanism can lead to enhanced representations especially in deeper layers of a network, where compounded adjustments can refine the feature space progressively.
\end{itemize}

\subsection{Gated Residual Connections}
Gated Residual Connections (GRU) extends the idea of \cite{srivastava2015highway} by enhancing the standard residual connections in the transformers architecture with a gating mechanism. This gating mechanism is used to control the flow of information effectively, allowing selective emphasis or suppression of features based on their relevance determined by the gating mechanism. We replace the standard residual connections that bypasses the multi-head attention modules and feed forward modules in encoder and decoder layers of the transformer architecture \cite{NIPS2017_3f5ee243} with our proposed Gated Residual Connections. 

\subsubsection{Gating mechanism}
Each GRC employs a gating mechanism that computes a gate vector $\mathbf{g}$ to control the contribution of sublayer outputs:
\begin{equation}
\mathbf{g} = \sigma(\mathbf{W}_g \mathbf{r} + \mathbf{b}_g)
\end{equation}
where $\sigma$ is the sigmoid function, $\mathbf{r}$ is the residual output and $ \mathbf{W}_g \in \mathbb{R}^{k \times k} $ and $ \mathbf{b}_g \in \mathbb{R}^k$ are the weights and biases of the gating layer.

\subsubsection{Application of gated residual connections}
The gate vector $\mathbf{g}$ scales the sublayer output $\mathbf{s}$ (output from an encoder/decoder layer) before it is added back to the input:
\begin{equation}
\mathbf{y} = \mathbf{r} + (\mathbf{g} \odot \mathbf{s})
\end{equation}
This selective scaling allows the model to dynamically adjust how much of each sublayer's output should influence subsequent layers. Refer Figure \ref{integration_img} to visually see the integration of Gated Residual Connections in the Transformer model introduced in \cite{NIPS2017_3f5ee243}.

\section{Experiments and evaluation}
\label{Experiments_and_Evaluation}
We systematically assess the performance of our newly proposed Evaluator Adjuster Unit and Gated Residual Connection across a spectrum of tasks in natural language processing (NLP). Some of our experiments are inspired from \cite{NIPS2017_3f5ee243} and \cite{su2021roformer}. We commence by exploring the individual and combined effects of these mechanisms on the sequence-to-sequence machine translation task as detailed in Section \ref{Machine_Translation}. Subsequently, in Section \ref{Pre-training_Language_Modeling}, we investigate their impact during the pre-training phase of BERT \cite{devlin-etal-2019-bert}, both separately and in conjunction. Following the pre-training, we fine-tune and evaluate these models on various downstream tasks derived from the GLUE Benchmarks \cite{wang-etal-2018-glue}, with results discussed in Section 4.3. Additionally, in Section 4.4, we train and assess our approaches using the Multi30K dataset \cite{elliott-etal-2016-multi30k} and verify that the improvements in model performance is not due to an increase in number of model parameters. All experiments were run on NVIDIA Tesla V100 GPU.

\subsection{Machine translation}
\label{Machine_Translation}
To assess the efficacy of our proposed enhancements in sequence-to-sequence language translation tasks, we conducted experiments using the well-established WMT 2014 English-German dataset \cite{bojar-EtAl:2014:W14-33}, which comprises approximately 4.5 million sentence pairs. These experiments aim to compare the performance of models enhanced with our Evaluator Adjuster Unit (EAU) and Gated Residual Connections (GRC) against the standard Transformer model \cite{NIPS2017_3f5ee243}.

\begin{table}
  \caption{BLEU scores \cite{papineni-etal-2002-bleu} comparison with baseline Transformer \cite{NIPS2017_3f5ee243} and its enhanced variants on the WMT 2014 English-to-German translation task \cite{bojar-EtAl:2014:W14-33}.}
  \label{Comparing_Baseline_T_and_T_enhanced_wmt14}
  \centering
  \begin{tabular}{ll}
    \toprule
    Model & BLEU score \\
    \midrule
    Baseline Tranformer \cite{NIPS2017_3f5ee243} & 26.61 \\
    Transformer with EAU & 26.69 \\
    Transformer with GRU & 26.77 \\
    Transformer with EAU and GRU & \textbf{26.79} \\
    \bottomrule
  \end{tabular}
\end{table}

All models, including the baseline Transformer \cite{NIPS2017_3f5ee243}, were trained under identical conditions to ensure a fair comparison. We configured each model with a maximum sequence length of $n = 512$ tokens and set the dimensions for keys, queries, and values at $k = 512$. The dimension of the feed-forward network in each transformer block was set to $f = 2048$, and a dropout rate of 0.1 was applied to prevent overfitting. Optimization was performed using the AdamW optimizer \cite{Loshchilov2017DecoupledWD}, with hyperparameters $\beta_1 = 0.9$, $\beta_2 = 0.98$ and weight decay of $0.01$. The learning rate was initialized at 0.001 with a warm-up period of 4000 steps, and label smoothing was implemented with a factor of 0.1.

The performance of each model variant was measured using BLEU scores \cite{papineni-etal-2002-bleu}, a standard metric for evaluating translations. The results, presented in Table \ref{Comparing_Baseline_T_and_T_enhanced_wmt14}, indicate that models incorporating the proposed EAU and GRC outperform the baseline Transformer model, demonstrating the effectiveness of these enhancements in improving translation quality.

\subsection{Pre-training language modeling}
\label{Pre-training_Language_Modeling}
In this experiment, we assess the efficacy of the Evaluator Adjuster Unit (EAU) and the Gated Residual Connection (GRU), both individually and in combination, for learning contextual representations. Using the Huggingface Transformers library (Apache License 2.0), we enhance the BERT \cite{devlin-etal-2019-bert} baseline model by integrating these components, utilizing the \texttt{bert-base-uncased} variant.

For the pre-training phase, we utilized the WikiText-103 dataset \cite{merity2016pointer} accessed from the Huggingface Datasets library, licensed under Apache License 2.0. This dataset was partitioned into 85\% for training and 15\% for validation. Pre-training was conducted using a batch size of 16 and a maximum sequence length of $n = 512$ across 100,000 steps. Optimization was performed using the AdamW optimizer \cite{Loshchilov2017DecoupledWD} with hyperparameters $\beta_1 = 0.9$, $\beta_2 = 0.98$, weight decay of $0.01$ and a learning rate of 5e-5. We evaluated the models using the masked language-modeling (MLM) loss as a metric.

\begin{figure}
  \centering
  \includegraphics[width=0.6\textwidth]{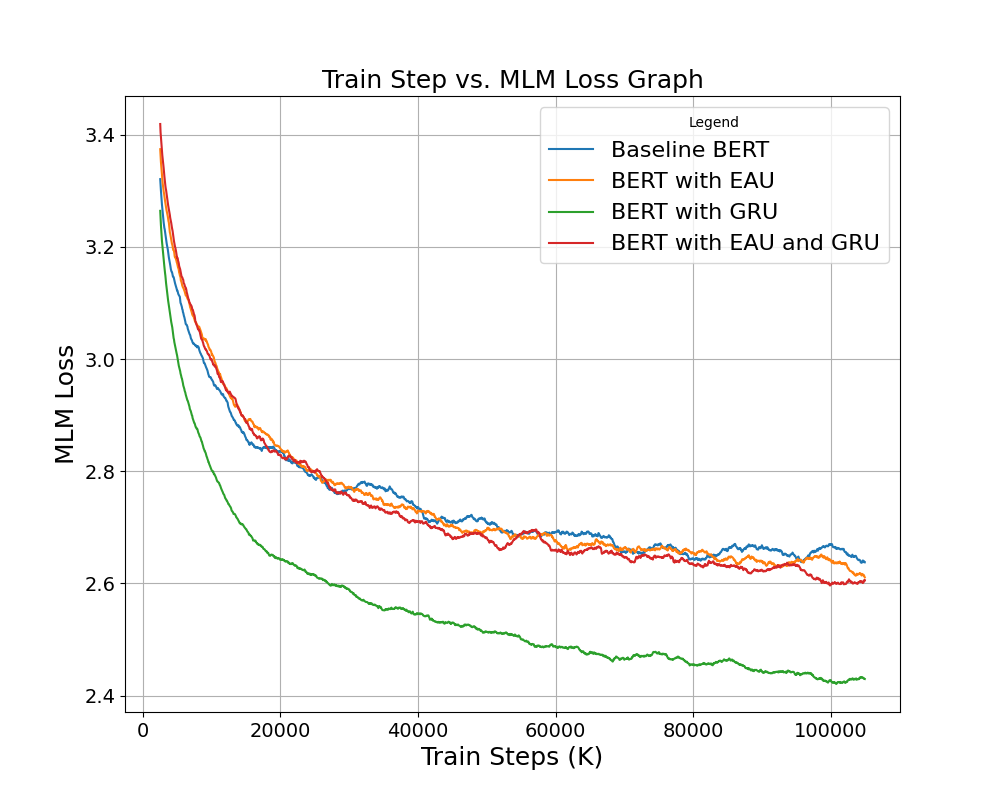}
  \caption{Masked Language Modeling loss for different models.}
  \label{mlm_loss}
\end{figure}

Figure \ref{mlm_loss} presents the MLM loss trajectories for various model configurations during the training process. The BERT model enhanced with EAU and the BERT model enhanced with both EAU and GRC exhibit slightly lower MLM loss compared to the Baseline BERT model. Finally, the BERT model solely enhanced with GRC demonstrates the fastest convergence rate among the tested configurations.

\subsection{Fine-tuning on GLUE tasks}
\label{Fine-tuning_on_GLUE_tasks}

Building on the pre-trained models described in Section \ref{Pre-training_Language_Modeling}, we proceed to fine-tune their weights on various GLUE tasks \cite{wang-etal-2018-glue} to evaluate their generalization capabilities across a range of downstream NLP tasks. Specifically, the models are fine-tuned on the Microsoft Research Paraphrase Corpus (MRPC) \cite{mrpc}, Recognizing Textual Entailment (RTE) \cite{rte}, Winograd NLI (WNLI) \cite{wnli}, The Stanford Sentiment Treebank (SST-2) \cite{sst-2}, The Corpus of Linguistic Acceptability (CoLA) \cite{cola}, Question NLI (QNLI) \cite{wang-etal-2018-glue}, and the Semantic Textual Similarity Benchmark (STS-B) \cite{sts-b}.

Utilizing the Huggingface Transformers library (Apache License 2.0), fine-tuning is conducted on these downstream tasks with a batch size of 32 and a maximum sequence length of $n = 512$. Each task is fine-tuned for three epochs, with the exception of CoLA, which undergoes five epochs, and RTE, which extends to eight epochs. Optimization was performed using the AdamW optimizer \cite{Loshchilov2017DecoupledWD} with hyperparameters $\beta_1 = 0.9$, $\beta_2 = 0.999$, weight decay of $0.01$ and a learning rate of 2e-5.

The evaluation outcomes, along with the specific metrics used for each task, are detailed in Table \ref{Comparing_Baseline_BERT_and_BERT_fine_tuning}. Following the methodology of \cite{devlin-etal-2019-bert}, we present the best-averaged results from the validation sets. As we can see from the results, our proposed models perform better than baseline BERT with good improvements on many GLUE tasks.

\begin{table}
  \caption{Comparing baseline BERT and BERT enhanced with our approach after fine tuning on downstream GLUE tasks.}
  \label{Comparing_Baseline_BERT_and_BERT_fine_tuning}
  \centering
  \begin{tabular}{llllll}
    \toprule
    GLUE Task & Evaluation  & BERT & BERT with & BERT with & BERT with \\
              & Metric &      & EAU       & GRU       & EAU and GRU \\
    \midrule
    MRPC  & Accuracy(\%) & 73.04 & \textbf{81.86} & 78.92 & 79.41 \\
    RTE   & Accuracy(\%) & 57.04 & 58.84 & 57.40 & \textbf{62.45} \\
    WNLI  & Accuracy(\%) & 52.11 & 56.30 & 46.48 & \textbf{56.34} \\
    SST-2 & Accuracy(\%) & 89.79 & 88.88 & \textbf{89.91} & 89.68 \\
    QNLI  & Accuracy(\%)   & \textbf{86.31}  & 86.00       & 86.06  & 86.02 \\
    CoLA  & Matthew's Corr & 0.450 & 0.452 & \textbf{0.464} & 0.451 \\
    STS-B & Pearson-       & 0.833 & 0.821 & \textbf{0.840} & 0.835 \\
          & Spearman Corr  &        &        &        &       \\
    \bottomrule
  \end{tabular}
\end{table}

\subsection{Assessing model improvements beyond parameter increases}
\label{Assessing_Model_Improvements_Beyond_Parameter_Increases"}

\begin{table}
\caption{Model complexity vs. performance (
$l = $number of encoder, decoder layers; $n = $ maximum sequence length; $k = $key, query, value dimension; $f = $feed forward dimension)
}
  \label{Model_Complexity_vs_Performance}
  \centering
\begin{tabular}{lllll}
\toprule
                         & \multicolumn{2}{l}{}       & \multicolumn{2}{l}{EAU and GRC}        \\ 
                         & \multicolumn{2}{l}{Baseline Transformer}       & \multicolumn{2}{l}{integrated Transformer}        \\ 
                         \noalign{\vskip 2pt} 
                         \cline{2-5} 
                         \noalign{\vskip 2pt} 
 & \multicolumn{1}{l}{number of} & & \multicolumn{1}{l}{number of} & \\
 & \multicolumn{1}{l}{learnable} & & \multicolumn{1}{l}{learnable} & \\
Hyperparameters & \multicolumn{1}{l}{parameters} & BLEU & \multicolumn{1}{l}{parameters} & BLEU \\ 
\noalign{\vskip 2pt} 
\hline
\noalign{\vskip 2pt} 
$l = 3$; $n = 128$; $k = 256$; $f = 1024$ & \multicolumn{1}{l}{11,066,797}  & 40.540 & \multicolumn{1}{l}{13,239,085}  & 48.876  \\
\noalign{\vskip 2pt} 
\hline
\noalign{\vskip 2pt} 
$l = 2$; $n = 128$; $k = 256$; $f = 1024$ & \multicolumn{1}{l}{9,223,597}  & 39.432 & \multicolumn{1}{l}{10,671,789}  & 47.847  \\
\noalign{\vskip 2pt} 
\hline
\noalign{\vskip 2pt} 
$l = 2$; $n = 64$; $k = 128$; $f = 512$ & \multicolumn{1}{l}{3,698,221}  & 38.930  & \multicolumn{1}{l}{4,061,869}  & 47.483  \\
\bottomrule
\end{tabular}
\end{table}

To demonstrate that improvements in model performance are not merely attributable to an increase in the number of learnable parameters, we conducted a series of experiments with several model variants. We initially established a baseline using a standard transformer model, akin to the architecture described in \cite{NIPS2017_3f5ee243}. In parallel, we developed a variant of this model that integrates our proposed Evaluator Adjuster Unit (EAU) and Gated Residual Connections (GRC), maintaining identical hyperparameters to the baseline to ensure comparability.

While the EAU and GRC variant inherently possesses a higher count of learnable parameters, as detailed in Table \ref{Model_Complexity_vs_Performance}, we also crafted two additional versions of this enhanced model. These versions were designed with adjusted hyperparameters aimed at reducing the number of learnable parameters such that they are comparable and even less than the baseline model's learnable parameter count. The specific hyperparameter modifications and their effects on the models' learnable  parameter count are documented in Table \ref{Model_Complexity_vs_Performance}.

All models were trained and validated on the Multi30K English to German translation dataset \cite{elliott-etal-2016-multi30k}. Performance metrics, as shown in Table \ref{Model_Complexity_vs_Performance}, indicate that both the standard and parameter-reduced variants of the EAU and GRC integrated model outperform the baseline transformer model, which has a comparatively higher parameter count.

Additionally, to address potential concerns of overfitting in the baseline model, we implemented reduced-parameter versions of the baseline transformer by altering the same hyperparameters used for the EAU and GRC models. These lighter models were also trained and validated on the Multi30K dataset in a manner consistent with the previous experiments. The resulting BLEU scores, presented in Table \ref{Model_Complexity_vs_Performance}, reveal a decrease for the lighter vanilla models, confirming that the baseline transformer was not overfitting. This comprehensive approach substantiates the effectiveness of our EAU and GRC enhancements beyond mere parameter scaling. Note: All other hyperparameters are the same for all the models discussed in this experiment. A batch size of 128 was employed. Optimization was performed using the AdamW optimizer \cite{Loshchilov2017DecoupledWD} with hyperparameters $\beta_1 = 0.9$, $\beta_2 = 0.999$, weight decay of $0.01$ and a learning rate of 2e-5.

\section{Limitations}
\label{Limitations}
While our study demonstrates promising enhancements to transformer architectures through the integration of the Evaluator Adjuster Unit (EAU) and Gated Residual Connections (GRC), we acknowledge several limitations:
\begin{itemize}
    \item \textbf{Necessity for Retraining:} Despite the relative ease of integrating EAU and GRC into existing transformer models, these modifications require retraining the models from scratch. This process involves significant computational resources and time, particularly for large-scale models.
    \item \textbf{Inconsistent Performance in Vision Tasks:} Our modifications have shown substantial improvements in natural language processing tasks. However, analogous gains have not been observed in vision-related applications. Further research is required to adapt and optimize these enhancements for vision tasks, ensuring that the benefits of EAU and GRC can be universally applied across different modalities.
\end{itemize}
These limitations highlight areas for future research and development, especially their application in diverse domains beyond NLP.

\section{Conclusion}
\label{Conclusion}
In this work, we introduced two novel enhancements to the transformer architecture: the Evaluator Adjuster Unit (EAU) and the Gated Residual Connections (GRC). These components were designed to improve the transformer's ability to adapt its attention mechanisms and information flow dynamically, based on the context of the input. The EAUs provide context-dependent modulation of attention scores and the GRCs allow the capability to adaptively gate information. Through extensive experimentation, we demonstrated that both EAU and GRC significantly enhance the performance of transformers across a range of benchmark datasets in natural language processing. We encourage both researchers and practitioners in the field to explore the incorporation of these modules into their transformer architectures, especially the GRC unit as it offers a lightweight yet powerful enhancement.

{
\small

\bibliographystyle{plain}

\nocite{*} 



}

\end{document}